\documentclass[journal]{IEEEtran}
\usepackage[numbers]{natbib}
\usepackage{multicol}
\usepackage{times}
\usepackage{multirow}
\usepackage{lipsum}
\usepackage[american]{babel}
\usepackage{graphicx} 
\usepackage{algorithm}
\usepackage{amsmath}
\usepackage{algorithmicx}
\usepackage{bm}
\usepackage{xcolor}
\usepackage{enumitem}
\usepackage{comment}
\usepackage{algpseudocode}

\begin{document}

\title{A Robotic Hand Synthesis Process for General Tree Topologies}
\title{Enumeration, structural and dimensional synthesis of robotic hands: theory and implementation}

\author{Ali~Tamimi, Taher Deemyad,
        and~Alba~Perez-Gracia,~\IEEEmembership{Member,~IEEE}
\thanks{Ali Tamimi (email: ali.tamimi@wsu.edu) is from the Department of Electrical Engineering \& Computer Science, Washington State University, Pullman, WA. Taher Deemyad (email: deemtahe@isu.edu) and Alba Perez-Gracia (email: perealba@isu.edu) is from the Department of Mechanical Engineering, Idaho State University, Pocatello, ID.}
}

\markboth{}%
{Shell \MakeLowercase{\textit{et al.}}: Bare Demo of IEEEtran.cls for Journals}

\maketitle

\begin{abstract}
Designing robotic hands for specific tasks could help in the creation of optimized end-effectors for grasping and manipulation. However the systematic design of robotic hands for a simultaneous task of all fingertips presents many challenges. In this work the algorithms and implementation of an overall synthesis process are presented, which could be a first step towards a complete design tool for robotic end-effectors.

Results on type synthesis for a given task and number of fingers, solvability and dimensional synthesis for arbitrary topologies are integrated and used to design an implementation procedure. These results have been developed over the last years and are connected in this work.  The resulting solver is a tool that can aid in the creation of innovative robotic hands with arbitrary number of fingers and palms. Several examples of type synthesis, solvability calculations and dimensional synthesis are presented to evaluate the method.
\end{abstract}


\IEEEpeerreviewmaketitle

\section{Introduction}

\IEEEPARstart{R}{obotic} hands are a type of mechanical linkage for which a common set of links spans several kinematic chains. Among the variety of robotic end-of-arm tools (EOAT), those generally defined as robotic hands are considered suited not only for grasping but also for some dexterous manipulation. 

When considering the design and applications of robotic grasping, and manipulation of grasped objects, many aspects of robotics have to converge, including sensing, identification, physical interaction, and motion control, to name a few; notable results are being obtained in each these fields. One field that has not attracted so much attention is the systematic methodology for the physical embodiment of the robotic end effector; however, its effect on the successful completion of the task may be considerable. About ten years ago, the design process for robotic hands started receiving some attention \cite{Ceccarelli:2008,SterusandTurner:2011,CiocarlieAllen:2010} but, given the complexity of the task, current research focuses on understanding what may be needed from hands and on benchmarking \cite{Parmiggiani:2019}. See \cite{Bicchi:2019} for a comprehensive review.

The design of end-effector robotic tools has focused on three different strategies \cite{Mason:2010}, which tend to yield different designs: anthropomorphism, designing for grasping tasks, and designing for dexterous manipulation. The anthropomorphic hands are more limited in their design options, but they are considered a straightforward solution for human environment and human manipulation task mapping \cite{Martin:2014,Mattar:2013,Zhu:2018,Choi:2019,Asfour:2019}. It is worth noting that these hands may be implemented using traditional joints or more recently using soft robotics \cite{Wang:2018} or other techniques \cite{Ren:2019}, but in any case, they share similar underlying kinematics. For a review of the design efforts of anthropomorphic hands, see \cite{Carrozza:2014}.

Hand design, when oriented to grasping tasks, yields usually simpler or under-actuated hands; new efforts are devoted to obtaining under-actuated or simple hands with variable degrees of dexterity \cite{Dollar:2014,Dollar:2018,Salisbury:2014,Ciocarlie:2018}. 

Hands for in-hand manipulation tend to be more complex either in structure, joint type or fingertip type, especially if a wide range of manipulation actions are targeted \cite{Salisbury:2020}; however, most of them are anthropomorphic in design 
\cite{Vamvoudakis:2019}. 

A task-based, systematic design process for robot hand kinematics, needs to consider the enumeration of topologies or structural synthesis, and the dimensioning of the selected topologies -the dimensional synthesis- followed by a stage of detailed design and implementation. 

The literature in type or structural synthesis is vast, especially for linkages with closed loops, which present more significant challenges in their classification. Type or structural synthesis is based on subgroups of motion, following \cite{Herve:1978}, and on the use of screw theory, such as \cite{KongGosselin:2004}, among others \cite{Zeng2012}, combined with graph theory for the enumeration and classification. Most of the current methods are based on defining subgroups of motion or subspaces of potential velocities for the system. A task-based approach for the structural synthesis needs to take into account the shape of the desired workspace, or kinematic task. Pucheta \cite{Pucheta2013} applied a graph theory-based method and precision position method consecutively for planar linkages and multiple kinematic tasks. Results on the structural synthesis of hands based on mobility while grasping an object has been studied in \cite{Tsai:2002,SalisburyRoth:1983}, and more recently in \cite{Ozgur:2014}. Our research in the type synthesis for structural hands can be found in \cite{tamimi2018structural}, where some of the ideas further developed in section IV of this article were outlined.

For the dimensional synthesis stage, most of the research has focused on the design of individual, underactuated fingers; see \cite{Robson:2014,Ceccarelli:2015,Alici:2019} for a traditional approach. The first tool for the systematic dimensional synthesis of complete multi-fingered robotic hands for given manipulation tasks, up to the authors' knowledge, was developed by Simo-Serra et al. \cite{Simo-Serra:2011}. This tool allows us to design multi-fingered robotic hands with a set of common wrist joints and a palm branching in the different number of fingers, see also \cite{SimoSerra:2012,SimoSerra:2014} for its theoretical development.

In this work, we present a complete design methodology for arbitrary robotic hands that includes topology enumeration and the corresponding arrays defining the topology,  structural synthesis for an input task, and dimensional synthesis for hands that can present several splitting stages. Theoretical aspects, algorithmic implementation, and computational aspects are included. The aim is to integrate these into a design tool to help in the creation of robotic hands tailored to specific applications.

\section{Tree Topologies}
\IEEEPARstart{A} tree topology for a kinematic chain has a set of common joints spanning several chains, possibly in several stages, and ending in multiple end-effectors \cite{Selig:2004}. A \emph{branch} of the hand is defined as a serial chain connecting the root node to one of the end-effectors, and a \emph{palm} is a link that is ternary or above.  The tree topology is represented as rooted in a tree graph; the approach of Tsai \cite{Tsai:2001} is followed, with the root vertex being fixed with respect to a reference system. 

A multi-fingered hand is defined as a kinematic chain with several common joints - the wrist, which is a fundamental part of the hand manipulation- spanning several branches, possibly in several stages. At the end of each branch are the end-effectors, the fingertips.  They are the main elements whose motion or contact with the environment is being defined by the task; this can be generalized to consider other intermediate vertices of the topology. Open hands, that is, hands not holding an object, are represented as kinematic chains with a tree or hybrid topology. For our synthesis formulation, the internal loops in the hand structure are removed using a reduction process \cite{SimoSerra:2014}, to obtain a tree topology with intermediate links that are ternary or above.

\begin{figure}[h!]
\centering
\includegraphics[width=2.8in]{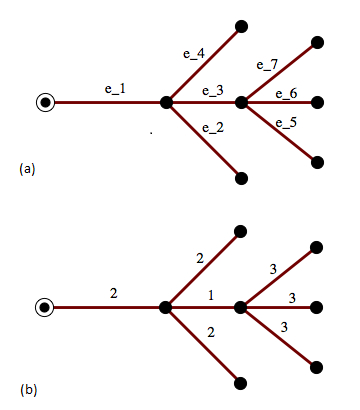}
\includegraphics[width=3.2in]{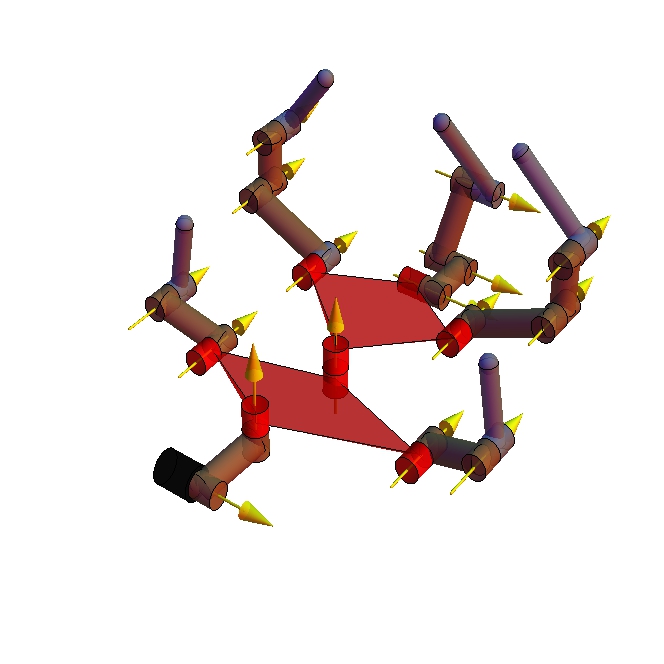}
\caption{ A five-fingered, two-palm hand topology. (a) indicates the numbering of the edges and (b) indicates the number of joints for each edge. Below: A kinematic sketch of the hand.}
\label{fig:hand1}
\end{figure}

Tree topologies are denoted as $SC-(B_1, B_2,\ldots, B_b)$, where $SC$ is a serial kinematic chain representing the initial common joints and the dash indicates a branching or splitting, with the branches adjacent to $SC$ contained in the parenthesis, each branch $B_i$ characterized by its type and number of joints. Figure \ref{fig:hand1} shows the compacted and possibly reduced graph for a $2R-(2R,R-(3R,3R,3R),2R)$, or  $2-(2,1-(3,3,3),2)$ chain if we drop the $R$ in the case of all revolute joints. This hand has three branches, one of them branching again on three additional branches, for five end-effectors or fingertips. The root vertex is indicated with a double circle. While most current robotic hands have a single splitting stage spanning several fingers, this can be generalized for greater adaptation to different applications by using hands with topologies such as the one presented in Figure \ref{fig:hand1}.

A tree topology is represented by two arrays, which capture incidence and adjacency properties as well as information on the edges.  Assume a numbering of the graph edges to define a \emph{parent-pointer array} and a \emph{joint array}. The length of both arrays is equal to the number of edges of the tree graph after the reduction process is applied, that is, each edge, and each entry of the arrays, will correspond to a serial chain of the robotic hand. 

The parent-pointer array implements the parent-pointer representation, where each element takes the value of the previous edge, the first edge being usually the one incident at the root vertex. The edges incident at the root vertex have no parent and they take the value zero. Each element of the joint array contains the number and type of joints for each edge. If we are limited to revolute joints, then the joint array element will equal the number of joints for that edge. As an example, for the tree topology shown in Figure \ref{fig:hand1}, the parent-pointer array and joint array are defined as $p=\{0,1,1,1,3,3,3\}$ and $j=\{2,2,1,2,3,3,3\}$ for the given numbering of the edges.

\section{Kinematic Synthesis}
\IEEEPARstart{K}{inematic}  synthesis, the process of creating a mechanical system for a given motion task, can be used to select and size a topology as a candidate hand design. The synthesis process for robotic hands has four main steps that are detailed below: task definition, type or structural synthesis, solvability calculations, and dimensional synthesis. After this process, we obtain a set of joints, connectivity, and relative position along a chain. Further steps of ranking, optimization and detailed design will be necessary to implement the candidates into functioning hands. 

\subsection{Task Definition}
\begin{figure}[h]
\centering
\includegraphics[width=3in]{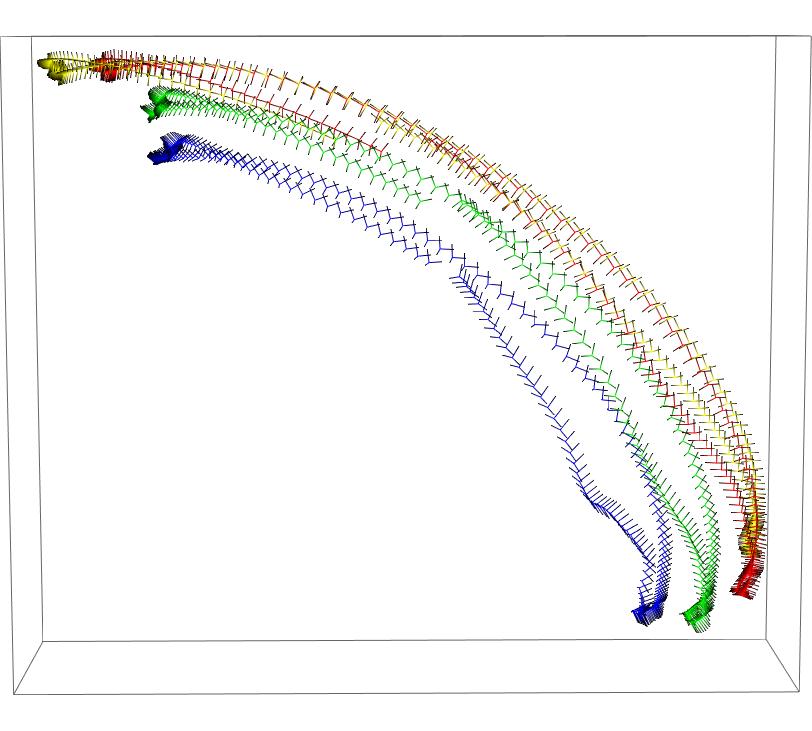}
\caption{ A motion task for a four-fingered hand, obtained using human hand motion capture.}
\label{fig:taskExample4}
\end{figure}

The task is the desired motion of the hand's elements whose interaction with the environment is of interest. For a multi-fingered hand, a \emph{simultaneous} motion of all fingertips or surface contacts, which could be any limb of the hand, is to be defined. For each fingertip or contact limb, a set of positions is defined as location plus orientation. Figure \ref{fig:taskExample4} shows a trajectory task for a hand with four fingertips.

\subsection{Type Synthesis}
Type synthesis, or structural synthesis, is the enumeration, selection and ranking of the kinematic chain topologies to be used as candidate designs. In the case of a robotic hand, it implies the selection or calculation of the number of fingers, the number of joints at the wrist, the number of splits or branchings, and the number of joints for the serial chain making each branch, as well as the type of joints to be used. This could be a prior selection by the designer, or it could be calculated based on the task. Having an automatic type synthesis stage helps the designer exploring the sometimes vast field of possible solutions and identify trends in the candidate topologies. 
In \cite{tamimi2018structural}, a structural synthesis, isomorphism-free enumeration method that combines the solvability for rigid-body guidance with grasping and manipulation metrics is developed and illustrated with validation examples.

\subsection{Solvability}
In the case of simultaneous tasks of all fingertips, solvability is defined as the ability of all combinations of fingers to perform their related tasks, and it is a condition that needs to be checked to be able to do the dimensional synthesis. It consists of checking the maximum number of positions that do not overconstrain each root-to-end-effector(s) subgraph, and ensuring that each subgraph is less constrained than the overall graph. In this calculation, the subgraphs obtained by moving the root to each end-effector need to be included to account for relative motion between fingertips.

\subsection{Dimensional Synthesis}
In the dimensional synthesis stage, the joint axes' positions are calculated for the selected solvable topology and the desired kinematic task. There are many techniques to state and solve the dimensional synthesis equations; regardless of the formulation, the output is the joint axes' position at a reference configuration. This output is equivalent to the set of parameters defining the relative location and orientation between adjacent joints. The kinematic solution can then be used for the detailed design of the hand.

\section{Type Synthesis and Enumeration}
\IEEEPARstart{G}{iven} a simultaneous motion task for all fingertips, it is essential to know how many, and what hand topologies are suited for the task. The number of candidate hand topologies of a specific type is usually very high and unbounded in some cases \cite{Makhal:2014}. This number of suited topologies can be reduced if some additional constraints are added. At the end of the process, one or a few of these topologies will be selected for performing the additional design steps. The approach taken here is different from previous research, such as \cite{Tsai:2002}, and it is based on free finger motion.

The conditions for considering a topology for the task are at least to have the same number of end-effectors as the task and solvable according to the criteria defined in \cite{SimoSerra:2014,Makhal:2014}. The set of suited topologies can be ranked according to other criteria, such as the number of edges, number of splits, and the number of joints per edge.

\subsection{Candidate topology search}
A search method and its algorithmic implementation are presented here to find solvable topologies for a defined task. The task is assumed to be a general subset in the $SE(3)$ group of rigid motion and its derivatives, and the goal is to find all topologies that can be paired with the task for dimensional synthesis, given a set of user-defined restrictions.  

User-defined inputs are the number of positions of the task $m$, the number of end-effectors, or branches, $b$, and the total number of edges of the graph $e$. Remember that every edge corresponds to a serial chain. The output is the set of topologies that meet the solvability criteria subject to these conditions. 

The task-sizing formula in \cite{SimoSerra:2014} is applied in the first place to find all possible branch topologies for the given number of end-effectors. Start by calculating 
\begin{equation}
J=\sum\limits_{i=1}^e j_{i} = \frac{(m-1)*6*b}{(m+3)}
\label{eq:condType}
\end{equation}
where $J$ represents the total number of joints, and $j_i$ are the joints for the serial chain corresponding to the $i$-th edge of the topology. The number of joints per edge has to be between $1$ and $5$ for synthesis purposes, as a serial chain of length $6$ or higher does not impose any restriction on the motion.

The presented method includes three steps. 
First, all possible tree structures (parent-pointer arrays $p$) that meet the input criteria are found for the given number of branches and edges. In the algorithmic implementation, the parent-pointer array is filled up, starting at the root and sequentially according to the following rules:
\begin{itemize}
\item $p(1)=0$. The first edge is the root node and has no parent.
\item If $i$ is not an end-effector, $p(i)$ can accept any value between $p(i-1)$  to $i-1$. The values of parent pointer array are increasing ($p(i)\geq p(i-1)$). This condition helps to avoid adjacent branch isomorphism.
\item If $i$ is an end effector, $p(i)$ can accept any value between $p(i-1)$  to $e-b$, since the last $b$ edges are end-effectors and cannot be parents.
\end{itemize}
Second, for each structure found in the first step, construct all possible joint arrays which meet the input criteria. This implies writing all joint arrays with length equal to $e$ that satisfy Eq.~\ref{eq:condType} and with entries between $1$ and $5$.  Constructing isomorphic trees is avoided by proper index and value assignment.

Finally, after finding all possible joints arrays for each parent pointer array, check each topology's solvability, including parent pointer array and joint array. If it is solvable, add it to the result as a candidate topology. This method yields all non-isomorphic trees \cite{Jaggard:2011}  for the input parameters.

The algorithmic implementation of the structural synthesis for tree topologies is detailed in Algorithm~\ref{alg:search}.
\begin{algorithm}
\caption{Candidate Topology Search}\label{alg:search}
\begin{algorithmic}
\Procedure{$ParentPointerArrayFinder$}{$b,e$}
\Comment{b is total number of branches and e is total number of edges }
   \State $max\gets e-b$
   \State $arrays\gets \Call{makeArrays}{$e$}$
   \Comment{makeArrays make all possible arrays which have the 3 conditions which are explained above}

   \ForAll{$ arrays $}
        \If{$numberOfBranches(array)=b$}  
              \State $PossibleParentPointerArrays.Add(array)$
       \EndIf
   \EndFor
   \State \textbf{return} $PossibleParentPointerArrays$
\EndProcedure

\Procedure{$JointArrayFinder$}{$P,J,e$} \Comment{P is parent pointer array, J is total number of joints and e is total number of edges}

   \State $first\gets Smallest Number With (e) Digits Without 0$ 
   \Comment{111..11 (with length of e) }
   \State $last\gets Largest Number With (e) Digits$
   \Comment{555..55 (with length of e- don't need digits larger than 5) }
   \For{$i\gets first, last$}
       \If{$digitsSum(i)=J$}  
             \State $check\gets true$
              \ForAll{$e\ IN\ end effectors$}  \Comment{the last e-b digits are end effectors}
              	 \If{$P[e]=P[e-1]\ and\ digit[e] < digit[e-1]$} 
              	 	\State $check\gets false$
		 \EndIf
	      \EndFor 
              	 \If{$check=true$} 
              	 	 \State $PossibleJointArrays.Add(i)$
		 \EndIf
       \EndIf
   \EndFor
   \State \textbf{return} $PossibleJointArrays$
\EndProcedure  
\Procedure{TopologySearch}{$P,b,e$}\Comment{P is parent pointer array, b is total number of branches and e is total number of edges}
   \State $parentpointer\_array \gets \Call{ParentPointerArrayFinder}{$b,e$}$
   \ForAll{$parentpointer\_array$}
        \State $joint\_array \gets \Call{JointArrayFinder}{$parentpointer\_array,J,e$}$
            \If{Solvable($joint\_array,parentpointer\_array)$}  
            \Comment{Solvability algorithm is explained in the following}
                \State $results.Add(joint\_array,parentpointer\_array)$
            \EndIf
   \EndFor
   \State \textbf{return} $results$
\EndProcedure
\end{algorithmic}
\end{algorithm}

\subsection{Type synthesis enumeration}

Even though this search may be unbounded, the reduced atlas can be created for a specific range on the number of end-effectors and precision positions.

Table~\ref{tab:candidateTopology} shows different values for the inputs and the number of candidate topologies that can be found. In this table, $m $ is the number of task positions for each fingertip, $b$ is the number of end-effectors, $e$ is the number of edges of the graph. The overall number of joints is calculated under \emph{Joints}, and the number of different joint arrays ($j$) and parent-pointer arrays ($p$) are calculated. The candidate topologies are the solvable combinations of joint arrays and parent-pointer arrays.

\begin{table}[h!]
\caption{Type synthesis results for selected inputs}
\label{tab:candidateTopology}
\begin{center}
 \begin{tabular}{ |c c c | c c c c|}
\hline
\multicolumn{3}{|c|}{\textbf{ INPUTS}} & \multicolumn{4}{|c|}{\textbf{OUTPUTS}} \\
\hline
\textbf{m} & \textbf{b} & \textbf{e} & \textbf{Joints} & \textbf{j} & \textbf{p} &\textbf{Candidate Topologies}  \\
\hline \hline
3 & 2 & 2& 4 & 2 & 1 &1  \\
\hline
3 & 2 & 3 & 4 & 2& 1 &2  \\
\hline
3 & 3 & 3 & 6 & 3 & 1 &1  \\
\hline
5 & 2 & 3 & 6 & 6 & 1 &4  \\
\hline
5 & 3 & 3 & 9& 5& 1 &1  \\
\hline
5 & 3 & 4 & 9 & 45 & 2 &9 \\
\hline
5 & 3 & 5 & 9 & 46 & 1 &19  \\
\hline
5 & 4 & 4 & 12 & 8 & 1 &1  \\
\hline
5 & 4 & 5 & 12 & 187 & 3 & 14  \\
\hline
5 & 4 & 6 & 12 & 478 & 3 & 72  \\
\hline
5 & 4 &7 & 12 & 206 & 1 & 47  \\
\hline
6 & 3 & 4 & 10& 58& 2 &4  \\
\hline
6 & 3 & 5 & 10& 76 & 1 &13  \\
\hline
9 & 4 & 4 & 16& 5& 1 &1  \\
\hline
9 & 4 & 5 & 16 & 250 & 3 &26  \\
\hline
9 & 4 & 6 & 16 & 1442 & 3 &237  \\
\hline
9 & 4 & 7 & 16& 1313& 1 &292  \\
\hline
13 & 2 & 3 & 9& 11 & 1 &6  \\
\hline
13 & 4 & 5 & 18 & 187 & 3 &4  \\
\hline
13 & 4 & 6 & 18 & 1645 & 3 &161  \\
\hline
13 & 4 & 7 & 18 & 2137 & 1 &233  \\
\hline
13 & 6 & 7 & 27 & 781 & 5 &2  \\
\hline
21 & 2 & 3 & 10 & 10 & 1 &10  \\
\hline
21 & 3 & 3 & 15 & 1 & 1 & 1  \\
\hline
21 & 3 & 4 & 15 & 45 & 2 & 24  \\
\hline
21 & 5 & 5 & 25 & 1 & 1 & 1  \\
\hline
21 & 5 & 6 & 25 & 168 & 4 & 57  \\
\hline
\end{tabular}
\end{center}
\end{table}

\begin{figure}[h]
\centering
\includegraphics[width=1.5in]{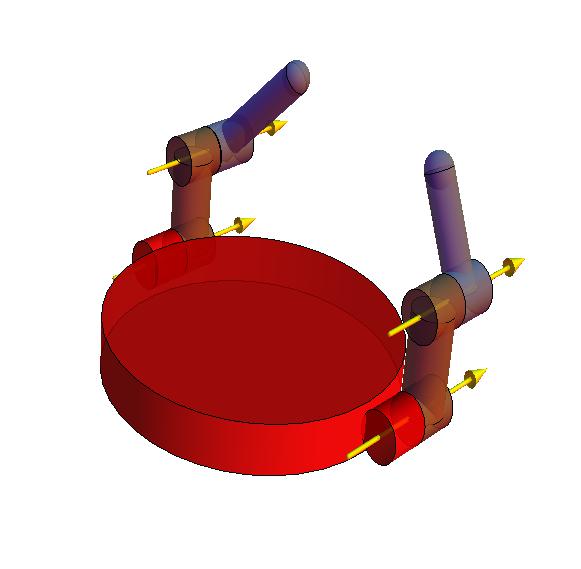}
\includegraphics[width=1.5in]{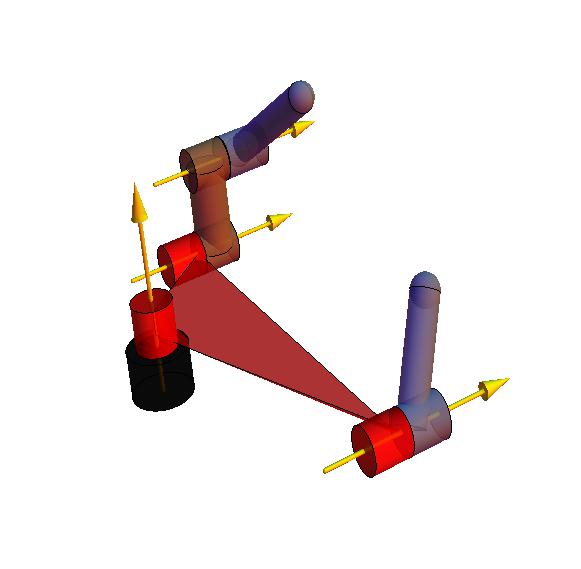}
\includegraphics[width=1.5in]{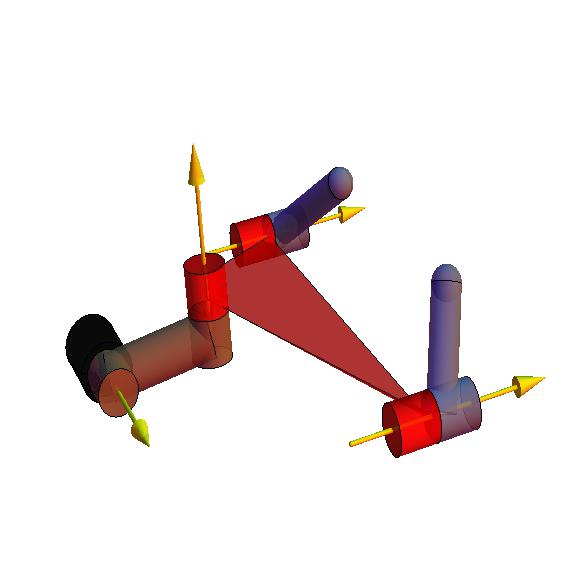}
\caption{ Kinematic sketch of all non-isomorphic candidate topologies with two fingertips and solvable for three precision positions. From left to right: 0-(2R,2R),R-(2R,R) and 2R-(R,R).}
\label{fig:sketches}
\end{figure}

To illustrate the results of Table~\ref{tab:candidateTopology}, Table~\ref{tab:examplesType} shows some of the candidate topologies that can be found using this method, where $p$ denotes the parent-pointer array and $j$ the joint array of the topology. Due to the high number of solvable candidate topologies, it is impossible to present them all in the table; however, the final number is presented in Table~\ref{tab:candidateTopology} for each example. Figure~\ref{fig:sketches} presents the three non-isomorphic topologies for two fingertips and three precision positions.

\begin{table}[h!]
\caption{Examples of type synthesis}
\label{tab:examplesType}
\begin{center}
\begin{tabular}{ |c|c| }
\hline
\textbf{Example 1} & \textbf{Example 2}\\
\hline
m=3 b=2 e=2\&3 & m=5 b=4 e=6 \\ 
 Topologies{:} & Some selected topologies{:} \\
 &  \\
 $p=(0,0)$ & $p=(0,0,1,1,2,2)$ \\
  $ j=(2,2)$ & $j=(1,1,2,3,2,3)$ \\
    &    \\
 $p=(0,1,1)$ & $p=(0,1,1,1,2,2)$ \\
   $j=(1,1,2)$& $j=(3,1,1,3,3,1)$ \\
    &    \\
  $p=(0,1,1)$ & $p=(0,1,1,2,2,2)$ \\
   $j=(2,1,1)$& $j=(3,2,1,3,2,1)$ \\
    &    \\
 & $p=(0,1,1,2,2,2)$  \\
 & $j=(2,2,2,2,2,2)$ \\ 
    & \\
   \hline
\textbf{Example 3} & \textbf{Example 4} \\ 
\hline 
  m=13 b=4 e=5 & m=21 b=5 e=6 \\ 
Some selected topologies{:} & Some selected topologies{:} \\
 & \\
 $p=(0,1,1,1,1)$ & $p=(0,0,0,0,1,1)$ \\
   $j=(2,4,4,4,4)$& $j=(2,5,5,5,4,4)$ \\
   &   \\
 $p=(0,1,1,1,1)$ & $p=(0,0,0,1,1,1)$ \\
   $j=(3,4,4,4,3)$& $j=(3,5,5,2,5,5)$ \\
   &    \\
  $p=(0,1,1,1,1)$ & $p=(0,0,1,1,1,1)$ \\
   $j=(4,3,4,4,3)$& $j=(4,5,3,5,3,5)$ \\
    &   \\
   $p=(0,1,1,1,1)$ & $p=(0,1,1,1,1,1)$ \\
   $j=(4,4,4,4,2)$& $j=(5,5,3,4,3,5)$ \\ 
    & \\
   \hline
\end{tabular}
\end{center}
\end{table}

\section{Solvability}
\IEEEPARstart{A}{} hand is defined as \emph{solvable} when It can be designed for a meaningful simultaneous task of all the fingertips or end-effectors, that is, a positive rational task with at least two positions. Because some fingers may be overconstrained while others are underconstrained for a given topology, solvability needs to be checked systematically for all root-to-end-effector subgraphs of the hand, including those obtained when changing the root vertex to one of the end-effectors. 

Equation~\ref{eq:position} calculates the number of positions for the exact kinematic synthesis of a tree topology. If the number of positions so obtained for all subtrees' kinematic task is greater or equal than the number of positions for the overall tree, the tree is solvable for kinematic synthesis. 

\begin{equation}
m= \frac{D_s^e . E - D_c^n . B }{D_{ee}^n . B - D_j^e . E }+1.
\label{eq:position}
\end{equation}

In this equation, $D_s^e $ is the vector containing the number of structural variables for each edge, $E $ is the vector of ones for the edges belonging to the subgraph, $D_c^n $ is the vector of possible extra constraints for each branch, $B $ is the vector of ones for the branches belonging to the subgraph, $D_{ee}^n $ is the vector of degrees of freedom for the motion of each end effector, and $D_j^e $ is the vector containing the number of joint variables for each edge. These vectors are calculated with the help of the root-to-end-effector path matrix of the graph.

The algorithmic implementation of the solvability condition has two steps, a first one to create all possible subgraphs and their corresponding arrays, and a second one to calculate the solvability for those subgraphs. Algorithm ~\ref{alg:solv} is implemented in order to calculate the solvability.

\begin{algorithm}[h]
\caption{Solvability}\label{alg:solv}
\begin{algorithmic}
\State $M\gets Number\_Of\_Position(Tree)$\Comment{equation ~\ref{eq:position}}
\ForAll{$SubTree$}
\State $m\gets Number\_Of\_Position(SubTree)$\Comment{equation ~\ref{eq:position}}
\If  {$M>m$} 
\State \textbf{return} $NOT Solvable$
\EndIf
\EndFor
\ForAll {end-effectors}
\Call{$RemoveCommonEdge$}{$Tree$}\Comment{Explain in Algorithm 3}
\State $newRoot\gets endeffector(i)$
\State
\Call{$Reconstruct$}{$Tree,newRoot$}\Comment{Explain in Algorithm 3}
\If  {only one end-effectors remain}
\State \textbf{return} $Solvable$
\EndIf
\EndFor
\end{algorithmic}
\end{algorithm}

The process of assigning a new parent-pointer array to the subtrees is shown in Algorithm~\ref{alg:pptSubt}. 

\begin{algorithm}
\caption{Change Parent Pointer Array}\label{alg:pptSubt}
\begin{algorithmic}
\Procedure{$RemoveCommonEdge$}{$Tree$}
\ForAll{ edge in edges}
\If  {edge is in all branches} 
\Call {$Remove$}{$edge$} \Comment{make value of edge in both ppt and joint arrays equal -1}
\EndIf
\EndFor
\EndProcedure
\Procedure{$Reconstruct$}{$Tree,newRoot$}
\Comment{change parent pointer for edges which are connected to the path between last root and current root}
\State $joint \gets newRoot$
\While {joint != zero}
\ForAll{edge in edges}
\If  {(\Call{$parent$}{$edge$}== \Call{$parent$}{$joint$})and(edge != joint)} 
\State  $parent(edge)\gets joint$
\EndIf
\EndFor
\State $joint\gets parent(joint)$
\EndWhile
\State
\Comment{change parent pointer for edges which are in the path between last root and current root (it means change the direction of path)}
\State
\State $p\gets 0$
\State $q\gets newRoot$
\ForAll {p in path}
\State $parent(q)\gets p$
\State $p\gets q$
\EndFor
\State $parent(q)\gets p$
\EndProcedure
\end{algorithmic}
\end{algorithm}

\begin{figure}[h]
\centering
\includegraphics[width=2.7in]{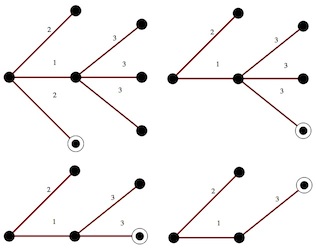}
\caption{ the trees which need to be checked for solvability.}
\label{fig:handSubG}
\end{figure}

\begin{figure}[h]
\centering
\includegraphics[width=3.5in]{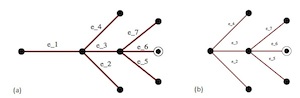}
\caption{ Parent-pointer array for subtrees:  a) Initial  b) after change (next root).}
\label{fig:handChangeRoot}
\end{figure}

As an example, for the tree topology shown in Figure \ref{fig:hand1}, solvability needs to be checked for the original tree, the following root-changing trees (Figure \ref{fig:handSubG}) and all of their subtrees.   Figure \ref{fig:handChangeRoot} shows the new parent-pointer representation assignment.

The process can be itemized as follows: 

\begin{itemize}
\item[a)] Remove the common edge and set parent pointer as zero for those edges, which are the common edge child. Figure \ref{fig:handSubG} shows the tree's previous root in the first tree and the tree after changing the root to the next root in the second one, and further steps in the other two graphs. The parent pointer array after this step changes from $ppt=\{0,1,1,1,3,3,3\}$ to $ppt=\{-1,0,0,0,3,3,3\}$. The value $-1$ means that the edge has been removed.

\item[b)] There is a path between previous root and new root. In this step, the parent-pointer value of the edges that are connected in this path is updated. In the example of Figure~\ref{fig:handChangeRoot}, the path includes edges $3$ and  $6$. In this step, the value of parent pointer for edges $2,4,5,7$ is changed. The parent-pointer array after this step changes from $ppt=\{-1,0,0,0,3,3,3\}$ to $ppt=\{-1,3,0,3,6,3,6\}$.

\item[c)] Finally, the parent-pointer value for the edges which are in the path is updated, by changing the value of parent pointer for edges $3$ and $6$.  The parent-pointer array after this step changes from $ppt=\{-1,3,0,3,6,3,6\}$ to $ppt=\{-1,3,6,3,6,0,6\}$.

\item[d)] After the previous step, the new tree is ready and $m$ can be calculated for all the combinations of the branches, as the algorithm shows.   Knowing the branch connectivity is needed for making the $[B]$ matrix, and the tree with $b$ branches has $2^b-1$ combinations of branches for the original root node; when switching the root node to each end-effector, that yields $2(2^b-1)-b$ different subtree combinations. These are defined by changing $1 \to 2^j$ to binary numbers using $j$ digits, $j=1, \ldots b$. Finally, all the needed matrices are available for calculating $m$ and comparing them to $M$.
\end{itemize}

\subsection{Solvability examples}

Table \ref{solvExamples} shows the results of the solvability checking algorithm for some hand topologies. For the cases in which the topology is solvable, the number of positions used for exact kinematic synthesis is returned. If the tree is not solvable, the overconstrained subtrees are identified.

\begin{table}[h!]
\caption{Examples of solvability calculation}
\label{solvExamples}
\begin{center}
\begin{tabular}{|lc|c||}
\hline
Topology & Solvability \\

\hline
3R-(2R,R-(3R,3R,3R,3R)) & Solvable  $m = 7$ \\
$p=(0,1,1,2,2,2)$ &  \\
$ j=(3,1,2,1,1,1)$ &  \\
  &  \\
 \parbox[c][2.4cm][c]{2.4cm}{ \includegraphics[scale=0.34]{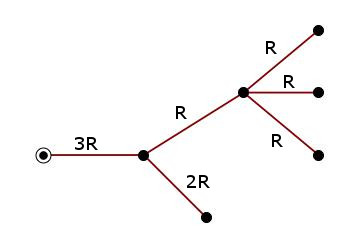} } &  \parbox[c][3.0cm][c]{3.0cm}{ \includegraphics[scale=0.19]{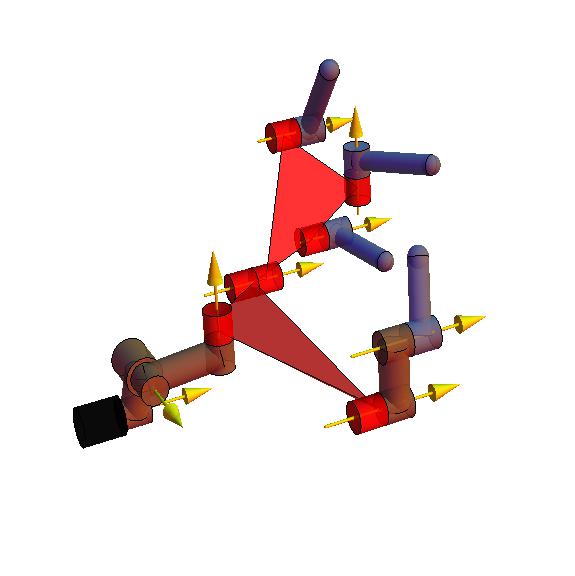} }\\
\hline
R-(2R-(3R-(R,R),3R-(R,R)),& Not Solvable \\
 2R-(3R-(R,R),3R-(R,R)))  & R-(R) overconstrained  \\
 $p=(0,1,1,2,2,3,3,4,4,5,5,$ & \\
 $6,6,7,7)$ &  \\
 $ j=(1,2,2,3,3,3,3,1,1,1,1,$ & \\
 $1,1,1,1)$ &  \\
  &  \\
\parbox[c][3.4cm][c]{3.4cm}{ \includegraphics[scale=0.32]{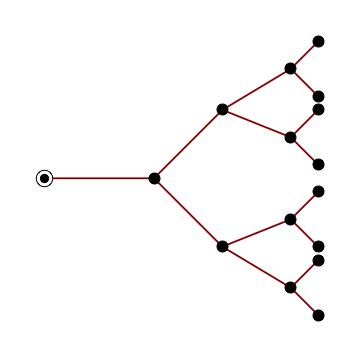} } & \parbox[c][3.0cm][c]{3.0cm}{ \includegraphics[scale=0.19]{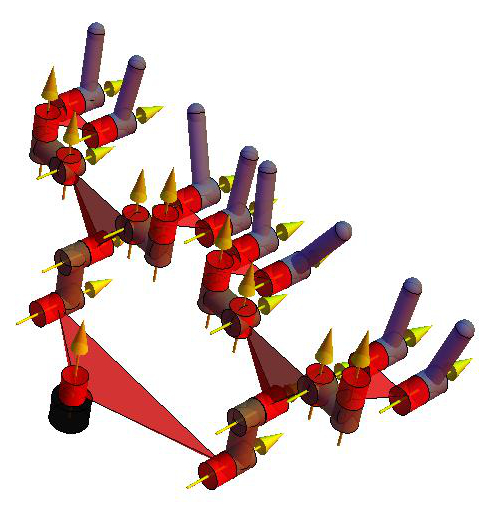} }\\
\hline
R-(R-(2R-(R,R-(R,R)),R-(R,R)), & Solvable $m = 3$ \\ 
 R-(R,R)) & \\
$p=(0,1,1,2,2,3,3,4,4,5,5,9,9)$ &  \\
 $ j=(1,1,1,2,1,1,1,1,1,1,1,1,1)$ &  \\
  &  \\

 \parbox[c][2.8cm][c]{2.8cm}{ \includegraphics[scale=0.35]{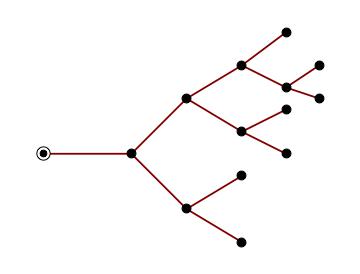} } &  \parbox[c][3.0cm][c]{3.0cm}{ \includegraphics[scale=0.19]{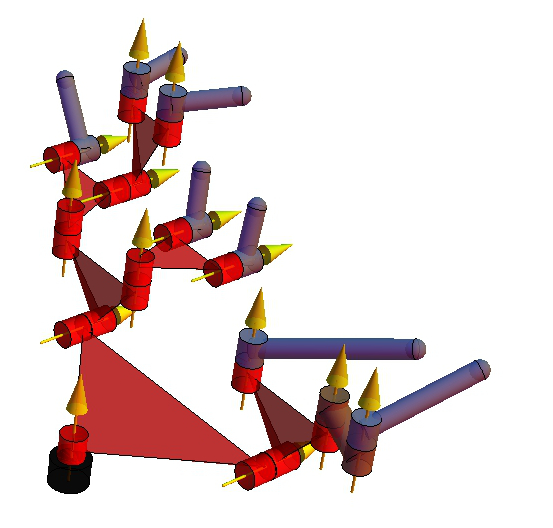} }\\
  & \\
 \hline
2R-(3R,R-(2R,2R,2R),3R) & Solvable $m = 5$ \\
$p=(0,1,1,1,3,3,3)$ &  \\
 $ j=(2,3,1,3,2,2,2)$ &  \\
  &  \\
 \parbox[c][3.0cm][c]{3.0cm}{ \includegraphics[scale=0.19]{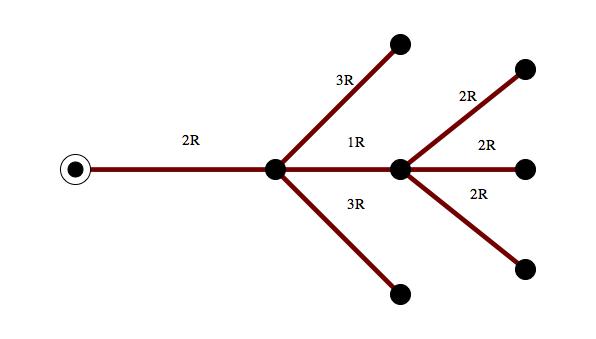} } &  \parbox[c][3.0cm][c]{3.0cm}{ \includegraphics[scale=0.19]{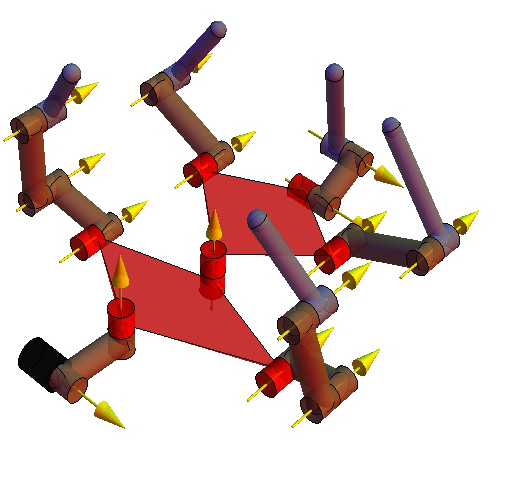} }\\
  & \\
\hline
\end{tabular}
\end{center}
\end{table}

\section{Structural Synthesis Algorithm for Free-Finger and Object-Contact Tasks}

The goal is to find all hand topologies paired with the task for dimensional synthesis, given a set of user-defined restrictions. User-defined inputs are the number of positions of the task $m$, the number of end-effectors or branches $b$, the range $[e_{\min},e_{\max}]$ for the total number of edges of the graph $e$, the types of allowed fingertips $\mathbf{c}$, and the desired mobility conditions. 

The overall mobility $M$ of the grasped object can be imposed as an input. Different mobility can be imposed at a given palm level, $M(T_{p_i})$, for in-palm manipulation or grasping, that can be different from $ M $ for specific sub-tasks of the task. 

The output is the set of topologies that (i) meet the solvability criterion subject to these requirements, and (ii) meet the constraints related to the mobility.

\subsection{Full-tree Mobility Conditions}
Given the tree of the hand $T$ and its mobility $M$, any root-to-end-effector subgraph $T_{\mathrm{sub}}$ must satisfy non-positive mobility $M'$ if joints are locked and mobility greater or equal than that of the overall tree.

\begin{equation}\label{eq:SubtreeM}
\forall T_{\mathrm{sub}} \in T \, :\qquad
\begin{cases}
M'(T_{\mathrm{sub}}) \leq 0 & \\
M(T_{\mathrm{sub}})  \geq M.& \\
\end{cases}
\end{equation}

\subsection{Variety}
Tischler and Hunt \cite{TischlerHunt:1995b} define the variety of a graph as the difference between its full mobility $M$ and the minimum mobility of a subgraph containing a loop or set of loops, $M_{min}$, that is, $V = M-M_{min}$.

For the reduced and compacted tree graphs of the hand, all the loops contain the vertex corresponding to the grasped object. Imposing that the graphs have a variety $V=0$ ensures that the object has the desired degrees of freedom and that the locked-joints mobility is non-positive. This condition is imposed by identifying and checking the subgraphs created along with the tree graph, starting at the root. Let the ternary or above vertices (palms) be labeled as $p_i$, and the subgraph starting at $p_i$ in which all previous edges and vertices have been eliminated be $T_{p_i}$. 

For a graph with variety $V=0$, the following condition is imposed:
\begin{equation}
M(T_{p_i}) \geq M, \quad i = 1,\ldots, p,
\end{equation}
where $p$ is the total number of palms in the hand.

For task consisting of sub-tasks that require different mobility at different palm levels, the palms $p_i$ are identified, and the required mobility of the subtree starting at the edges spanning from the palm is checked for the candidate tree topologies,
\begin{equation}
M(T_{p_i}) = M_{p_i}, \nonumber
M'(T_{p_i}) \leq 0.
\end{equation}

\subsection{Solvability Conditions}

The solvability criterion for the dimensional synthesis of a tree topology $T$ passing through a number of $m$ positions is the formula $m=s(T)$ proposed by Simo-Serra and Perez-Gracia \cite{SimoSerra:2014}, which also requires the analysis for each subtree $T_{\mathrm{sub}}$ of the graph. The tree is solvable iff

\begin{equation}
s(T_{\mathrm{sub}}) \ge m \qquad \forall T_{\mathrm{sub}} \in T
\end{equation}

\subsection{Algorithmic Implementation} \label{alg:Alg1}

\begin{enumerate}
  \item \textbf{Find all the possible topologies.}
  \newline 
{\bf Inputs:} number of positions (m), number of branches (b), number of edges (e)
\newline 
{\bf Outputs:}Parent Pointer Array and Joints Array.
  \begin{enumerate}
    \item \textbf{Find parent pointer array (p).} Parent pointer array
must have length of $e$ and $b$ branches. 
    \item \textbf{Find joint array.} For each parent pointer array in
step 1.1, construct all possible joint arrays which meet the input
criteria.
  \end{enumerate}
  
  \item \textbf{Solvability check.} For each pair of the parent pointer array
and joint array found in step 1, calculates the number of positions for the exact kinematic
synthesis. If the number of positions obtained for the kinematic
task of all subtrees is greater or equal than the number of
positions for the overall tree, the tree is solvable.
  \begin{enumerate}
    \item \textbf{Find all root to end effectors subgraphs.} A graph with $b$ branches has $2^b-1$ subgraphs. Calculate $m$  for all subgraphs and compare them with $m$ for the overall tree.
    \item \textbf{Remove common edges.} Common edges are the edges that are contained in all branches. In this step, an algorithm finds all common edges and removes them.
    \item \textbf{Change root to one of the end effectors.} When the root of the graph has changed, the value of the parent pointer array and the joint array should be updated. The algorithm updates them in two steps. There is a path between the previous root and the new root.
\begin{itemize}
\item  First, the parent-pointer value of the edges that are connected to this path is updated.
\item Second, the parent-pointer value for the edges which are in the path is updated.
\item Other edges that are not in the path or do not connect to the path do not need to be updated because the parents of them did not change.
\end{itemize}
    \item Iterate steps 2.1 to 2.3. This part will be stopped
when only two end-effectors remain.
  \end{enumerate}
  
  \item \textbf{Mobility Check.} The output of step 2 is the
possible topologies. In this step, the algorithm verifies that the
topology's mobility is equal to that defined as input when the grasping loops are created, adding the fingertip contact array $\bf{c}$ to the graph.
  \begin{enumerate}
    \item \textbf{Remove unused part and calculate mobility.} Since some
part of the rigid body may not participate in the grasping process; the algorithm removes them. To find the used part, the algorithm finds all the edges in the branches from root to the end-effectors that contribute to grasping. The other edges are unused, and the value of $-1$ is assigned to each corresponding element of parent pointer array and joint array. Then, calculate mobility for the resulting topology. If it equals the user input, it is one of the possible solutions.
    \item \textbf{Find Mobility for subgraphs.} Using the algorithm proposed in step 2.1, find all the root to end-effector subgraphs and calculate mobility (M) and locked joint mobility (M') for
them.
    \item \textbf{Remove common edges (Palms).} Using the algorithm proposed in step 2.2 remove palms.
    \item Calculate Mobility for the graph of part 3.3.
    \item Iterate step 3.3 and 3.4 until there is no common edge.
    \item \textbf{Internal checks.} If all the subgraphs fulfill the two following conditions, the topology is one of the solutions.
\begin{itemize}
\item  $M'_{subgraph} \leq 0$
\item  $M_{subgraph} \geq M$
\end{itemize}
  \end{enumerate}
\end{enumerate}

The algorithm is divided into three main steps. In Step 1, the
algorithm searches all possible topologies which satisfy user inputs. Then, Step 2 checks the solvability of candidate topologies and keeps only the topologies that are solvable.
Finally, the solvable candidates' mobility is computed in Step 3, and those topologies that satisfy the user inputs are presented as final answers. The method is described in \ref{alg:Alg1}.

\subsection{Results}\label{sec:Results}
Table \ref{tab:exp} shows a binary hand in which the calculations are detailed for the overall mobility and in-palm mobility for different palms along with the depth of the tree, removing first the wrist and then the depth-1 palm. For clarity, the solvability of this hand is calculated separately.
\begin{table}[t]
\scriptsize
  \centering
  \caption{Mobility calculations for a binary hand with four fingertips.}
  \begin{tabular}{|p{2.7 cm} | p{3.2 cm} | p{0.6 cm}|p{1.1 cm}|} \hline
Topology &Parameter & Symbol & Value\\
    \hline
 & Number of task positions & $m$   & $9$ \\
  &  Number of branches (fingertips) & $b$ & $4$ \\

  $\mathbf{p}=\{0,1,1,2,2,3,3\}$ &   number of edges & $e$ & $7$ \\
  $\mathbf{j}=\{2,1,1,2,4,2,4\}$ &  Type of fingertip contact & $\bf{c}\rm$ & $\{2,2,2,2\}$ \\
  & Mobility & $M$ & 6 \\
  &  locked-joints mobility & $M$ & -10 \\
  \hline
  Subgraph 1 & Number of branches (fingertips) & $b$ & $4$ \\
 Remove Wrist &  number of edges & $e$ & $6$ \\
 $\mathbf{p}=\{0,0,1,1,2,2\}$ &  Type of fingertip contact & $\bf{c}\rm$ & $\{2,2,2,2\}$ \\
 $\mathbf{j}=\{1,1,2,4,2,4\}$ & Mobility & $M$ & 4 \\
  &  locked-joints mobility & $M$ & -10 \\
    \hline
  Subgraph 2 &  Number of branches (fingertips) & $b$ & $2$ \\
 Remove Palm 1& number of edges & $e$ & $2$ \\
  $\mathbf{p}$=\{0,0\}&  Type of fingertip contact & $\bf{c}\rm$ & $\{2,2\}$ \\
  $\mathbf{j}$=\{2,4\} & Mobility & $M$ & 4 \\
  &  locked-joints mobility & $M$ & -2 \\
    \hline
  \end{tabular}
  \label{tab:exp}
\end{table}

For comparison, the input used in the type synthesis example of Tischler and Hunt \cite{TischlerHunt:1995b} is used here in the first example below.  For the second example, we compare the output to the results of Salisbury and Roth \cite {SalisburyRoth:1983} but using soft fingers instead of pointy fingers with friction. The number of positions for the synthesis is chosen so that the number of joints in the hand candidates is similar to those in the references used. The input values for both examples are shown in Table \ref{tab:inputs}.
\begin{table}
\scriptsize
  \centering
  \caption{Input values for the example}
  \begin{tabular}{|c|c|c|c|} \hline
Parameter & Symbol & Example 1 & Example 2\\
    \hline
Number of task positions & $m$   & $5$  & $9$\\
Number of branches (fingertips) & $b$ & $3$ & $3$ \\
Minimum and maximum number of edges & $e$ & $(2,4)$ & $(2,5)$\\
Type of fingertip contact & $\bf{c}\rm$ & $\{3,3,3\}$ & $\{2,2,2\}$\\
Mobility & $M$ & 6 & $\ge 6$\\
    \hline
  \end{tabular}
  \label{tab:inputs}
\end{table}

For the first example, the algorithm constructed 95 hand topologies, and 10 of them were solvable. Out of those 10, only three topologies fulfilled the mobility requirements, that is, having $M=6$ at the object with negative locked-joints mobility. The 3 topologies are shown in Table \ref{tab:resultsEx1}. Out of these topologies, one of them has a 1-DOF wrist, which means that it has in-palm mobility equal to 5. The no-wristed hand obtained is the same that was obtained in the example from \cite{TischlerHunt:1995b}.

\begin{table}
\scriptsize
  \centering
  \caption{Resulting topologies suited for the tasks of Examples 1 and 2}
  \begin{tabular}{|c|c|c|c|} \hline
Example &Parent-pointer array & Joint array & Tree graph \\
    \hline
 &$\{0,0,0\}$ & $\{3,3,3\}$   &  \parbox[c][2.2cm][c]{3.0cm}{ \includegraphics[scale=0.14]{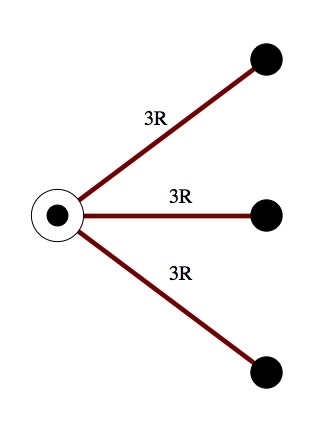} } \\
Example 1 & $\{0,0,1,1\}$ & $\{1,3,2,3\}$   &  \parbox[c][2.2cm][c]{3.0cm}{ \includegraphics[scale=0.21]{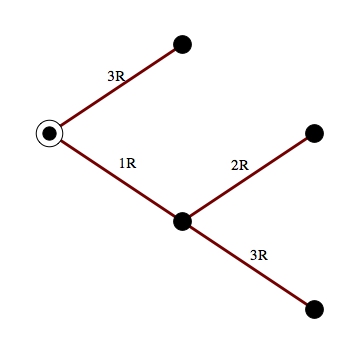} } \\
 & $\{0,1,1,1\}$ & $\{1,2,3,3\}$   &  \parbox[c][2.2cm][c]{3.0cm}{ \includegraphics[scale=0.21]{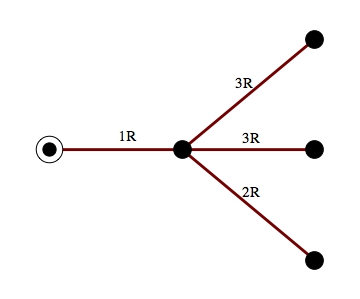} } \\
    \hline \hline
Example 2 &$\{0,0,0\}$ & $\{4,4,4\}$   &  \parbox[c][2.2cm][c]{3.0cm}{ \includegraphics[scale=0.14]{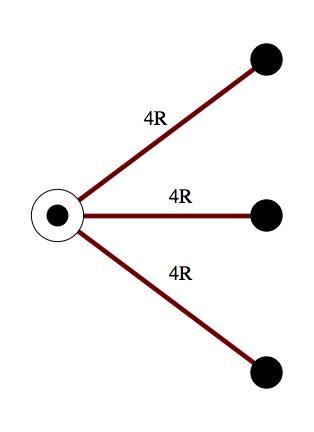} } \\
 \hline
  \end{tabular}
  \label{tab:resultsEx1}
\end{table}

For the second example, 295 topologies are compatible with the rigid-body guidance task, out of which 78 are solvable. However, only one topology, the one corresponding to three 4-DOF fingers and no wrist, has the required mobility $M = 6$ without being constrained by any subgraph, and negative locked-joints mobility. This topology corresponds to the solution chosen in \cite{SalisburyRoth:1983}. Several other topologies had the required overall mobility, but the additional constraint of having the same or higher in-palm mobility from any palm discarded those other topologies.

For the third case, we consider all topologies solvable for $m=5$ positions, with $b=3$ branches and $e=4$ edges, and pointy fingertips with friction. This yields the parent-pointer array $p=\{0,0,1,1\}$ shown in Table \ref{tab:exExtra}. It is required that the tree's overall mobility is $M = 6$ with locked-joints immobilizing the object. In addition, it is only required to have $M_1 = 3$ for the 2-finger palm.
The candidate topologies are shown in Table  \ref{tab:exExtra}, with only the last topology fulfilling all conditions. 

\begin{table}

  \centering
  \caption{Topology with different mobility at different palm levels}
  \begin{tabular}{|c|c|c|} \hline
Parent-pointer array &Joint array & Mobility \\\hline

  $\{0,0,1,1\}$ & $\{1,3,2,3\}$  & $M = 6$, $M_1=5$  \\
     & $\{2,3,1,3\}$  & $M = 6$, $M_1=4$  \\
          & $\{2,3,2,2\}$  & $M = 6$, $M_1=4$ \\
           & $\{3,3,1,2\}$  & $M = 6$, $M_1=3$ \\
 \hline
  \end{tabular}
  \label{tab:exExtra}
\end{table}

The results clearly show that the obtained hand topologies are general. Salisbury and Roth \cite{SalisburyRoth:1983} as well as Lee and Tsai \cite{Tsai:2002} procedures lead to hands with serial chain fingers and a unique palm without wrist. Ozgur methodology \cite{Ozgur:2014} leads to serial and complex (chains with loops) parallel hand topologies analogous to parallel robots. Tischler \textit{et al.} \cite{TischlerHunt:1995b} procedures have complex fingers with hybrid kinematic chains and produce topologies similar to those produced here for the case with a unique palm without wrist. Additionally, the tree topologies used here can be dimensioned through exact dimensional synthesis and, when connected to the grasped object, have serial, parallel, and hybrid topologies given more, or eventually new, design alternatives compared to those obtained in previous research.

The current algorithm also allows flexibility on where and when to define the mobility. The current implementation imposes the same or higher mobility at each palm as that of the overall hand, but
that can be modified to make some of the palms as grasping-only, for instance, while having different dexterity degrees depending on the palm and fingers involved.

\section{Dimensional Synthesis}
\IEEEPARstart{K}{iematic} dimensional synthesis is used to shape the new designs for robotic hands, able to grasp and/or manipulate in a given application. Dimensional synthesis has the candidate topology and the kinematic task as inputs. The kinematic task consists of a set of simultaneous displacements for each fingertip, as well as velocities and accelerations defined at some or all of those positions.

\subsection{Automatic forward kinematics equations}

For the design of robotic hands with arbitrary topologies, including multiple splitting stages, forward kinematics equations need to be automatically created from the tree topology and its associated arrays, identifying the common joints that will appear in the equations of several branches. The strategy to accomplish this is to divide the forward kinematics in serial chains -corresponding to graph edges-, branching points, and end-effector points. Three types of objects are defined as outlined below:

\begin{enumerate}
\item \emph{Chain}: a set of joint axes connected in series.  There are two different types of chains, those ending on an end effectors and those ending at a branching point. This second type is common to several branches; however, from the point of view of the object, they are generated equally.
\item \emph{Tip Contact Point (TCP)}: TCPs are created for each end-effector and then attached to the corresponding end-effector chain. A TCP can be attached to any link in the most general case, such as a palm link or intermediate finger link.
\item \emph{Splitter}: a vertex that spans more than one edge. Splitters are identified and created, and the chains spanning from each of them are attached to the splitter. If the splitter has a predecessor, then the splitter is attached to the common serial chain.
\end {enumerate}

This sequential process occurs until design equations are created for each chain from the root node to the end-effector node. The first step is generating the end effector chains.  For each end effector chain, generate and attach a TCP. For a single-branch topology (b=1), the process is done. For multi-finger topologies, Splitters are generated for each common joint, and chains are attached to them, from end-effector to root. Finally, the tree is attached to a first Splitter (\emph{sp0}) for topologies with no wrist or the serial chain of the first common edge in case of a wristed hand.  Algorithm \ref{attach} shows this process.

\begin{algorithm}[t]
\caption{Automatic FK object attachment}\label{attach}
\begin{algorithmic}
\If  {(no\_general\_wrist) } 
\State sp0=Create Splitter.
\EndIf
\State $endEffectors\gets\Call{FindEndEffectors}{$parentpointer\_array$}$
\Comment{endEffectors are those joints which are not pointed in parentpointer\_array}
\ForAll {e IN endEffectors}
\State Create chain.
\State attach TCP to e.
\If  {(no\_general\_wrist) and (parentpointer\_array[e]=0)} 
\State attach e to sp0.
\EndIf
\EndFor
\State $common\_joints\gets\Call{find\_palms}{$parentpointer\_array$}$
\Comment{palms or common joints are those joints which are attached by more than 1 joint.}
\ForAll {common\_joints}
\State Create chain.
\State Create Splitter.
\EndFor
\State $groups\gets\Call{makeGroup}{$parentpointer\_array$}$
\Comment{group[i] contains those joints which are attached to common\_joint[i]}
\For {$i\gets \#groups,1,step (-1)$}
\ForAll { joints IN group[i]}
\State attach joints to Splitter[i]
\EndFor
\State attach Splitter[i] to common\_joint[i]
\EndFor
\ForAll {cj IN common\_joints}
\If  {(no\_general\_wrist) and (parentpointer\_array[cj]=0)} 
\State attach cj to sp0.
\EndIf
\EndFor

\If  {b=1} \Comment{topology is single branch }
\State \Call{send}{$endeffector[1]$}
\EndIf
\If  {no\_general\_wrist} \Comment{topology is without wrist}
\State \Call{send}{$sp0$}
\EndIf
\If  {NOT(no\_general\_wrist)} \Comment{topology is with wrist}
\State \Call{send}{$common\_joint[1]$}
\EndIf
\end{algorithmic}
\end{algorithm}

\subsection{Exact synthesis}
Exact dimensional synthesis has been explored in \cite{SimoSerra:2014}.  The approach followed to create dimensional synthesis equations consists of equating the forward kinematics of each root-to-fingertip branch in hand to the set of positions defined for the fingertip.  Given a set of $m_p$ task positions $\hat{P}^i_{k}$, $k=1\ldots m_p$ for each end-effector (denoted by superscript $i$), $m_v$ task velocities ${\sf V}^i_r$ for each end-effector $i$, $r=1 \ldots m_v$, and $m_a$ task accelerations ${\sf A}^i_s$ for each end-effector $i$, $s=1 \ldots m_v$, where $m=m_p+m_v+m_a$,  design equations are created.  Compute the relative displacements from a selected reference position, usually position $1$, and equate the relative forward kinematics to those relative positions $\hat{P}^i_{1k}$. The twist of each end effector ${\sf V}^i_r$ is equated to the linear combination of twists for each joint axes, and similarly for the acceleration of the end effectors. The joint axes' Plucker coordinates appear explicitly in the forward kinematics when these are computed as the product of exponentials for relative displacements, and linearly in the velocity and acceleration equations. For a hand with $b$ fingertips, this yields $b$ sets of equations that are to be solved simultaneously,


\begin{align}
&\hat{P}^i_{1k} = 
\prod_{j \in \{B_i\}}    e^{\frac{\Delta\hat{\theta}_j^k}{2}{\sf S}_j},
                \nonumber \\
& {\sf V}^i_t = \sum_{j \in \{B_i\}} {\sf S}_j^t \dot\theta_j^t, \nonumber \\
& {\sf A}^i_r = \sum_{j \in \{B_i\}} {\sf S}_j^r \ddot\theta_j^r + \sum_{j,h \in \{B_i \}} \dot\theta_j^r \dot\theta_h^r [{\sf S}_j^r,{\sf S}_h^r ], \nonumber \\ 
&		\qquad 
                 i=1,\ldots,b;\quad k=2,\ldots,m_p; \quad t \in \{T_i\}; \quad r \in \{R_i\},
\label{eq:treedesign}
\end{align}

\noindent where the number of end-effectors, or branches as root-to-fingertip chains, is indicated by $b$,  $m_p$ is the number of exact positions, and $\{B_i\}$ is the set of ordered indices of the joints belonging to branch $i$, which can be obtained from the graph matrices. The set of ordered indices $\{T_i\}$ and $\{R_i\}$ correspond to positions where twists and accelerations have been defined, for each branch $i$. Notice that some of the joints will be common to several branches. The joint axes at the reference configuration are denoted as $\hbox{$\sf S$}_j$, and the joint axes at the configuration given by position $k$ are denoted as $\hbox{$\sf S$}_j^k$.

This yields a total of $6(m_p-1+m_v+m_a)b$ independent equations to be simultaneously solved.  The method has been applied to simultaneous rigid-body motion tasks for all fingertips \cite{Simo-Serra:2011}, defined by a finite set of positions, and to simultaneous fingertip tasks defined by a finite set of displacements and associated twists. For most topologies, this method yields many potential designs. 



\subsection{Multiple velocity synthesis or constrained-motion synthesis}
For tasks aiming to define a free trajectory for each fingertip, a finite set of positions, with a single twist vector defining the velocity for each position, and possibly a single acceleration 6D vector, gives a full characterization of the task.  However, for tasks constrained by the contact between the fingertip and an object, the definition of the allowed subspace of velocities at each point can be used to ensure the desired behavior for some grasping actions such as finger sliding or finger rolling, for a suited hand topology. Notice that the velocities must always be defined at a given position of the end-effector.

Consider the desired angular velocity of the end-effector and linear velocity of the end-effector frame's origin at a given position and calculate the fixed-frame six-dimensional twist. In this twist, the point velocity is calculated at the origin, so that it would yield the desired linear velocity at the origin of the end-effector frame. 

A constrained motion given by a contact is defined, at a given position, as a subspace of wrenches, $W$, whose magnitudes can be as high as needed. The subspace of reciprocal twists, $ V $, defines the allowed motion's potential directions at that position. 

A fingertip in contact with a surface can be kinematically modeled using one of the standard fingertip joints, see, for instance, \cite{Mason:2001}, such as pointy fingers or soft fingers, which are defined by their degrees of freedom and friction cone if applicable. For a general case, the subspace's dimension of reciprocal twists can be made to coincide with the mobility of the parallel mechanism formed when the hand is in contact with an object (defined by $n$ links and $j$ joints of $f_i$ degrees of freedom each), 
\begin{equation}
\dim(V) = 6(n-1)-\sum_{i=1}^j(6-f_i).
\label{eq:mob}
\end{equation}

Using this method, a hand can be synthesized for a desired $m$-dimensional subspace of twists at each precision position, just by defining a set of $m$ independent twists at the same position. 

\begin{figure}[h]
\centering
\includegraphics[width=2in]{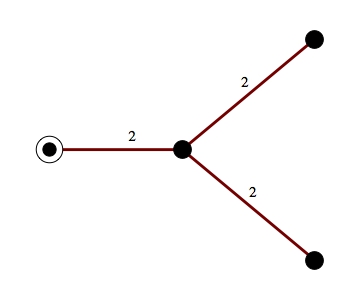}
\includegraphics[width=2.5in]{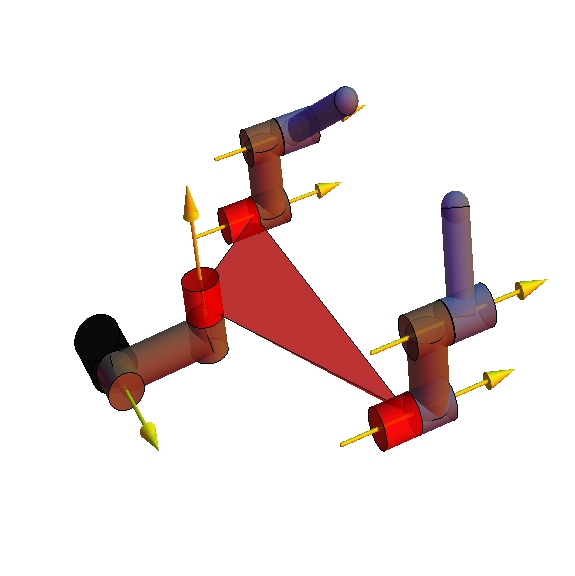}
\caption{ The 2-(2,2) hand: Up, topology; down, kinematic sketch.}
\label{fig:222}
\end{figure}

As an example, consider a hand with a $2-(2,2)$ topology, with two fingers and soft finger joints at the fingertip. Figure \ref{fig:222} shows the graph and kinematic sketch of the $1-(4,4)$ hand. This hand has two revolute joints at the wrist and two revolute joints at each of the two fingers. 

This topology is solvable for a total of $m=5$ precision positions. Define a task with $m_p=3$, with one position having two specified twists each ($m_v=2$), and the rest of positions having no specified velocities. If this is a task in which both fingers contact an object, and assuming a general grasp, the $2-(2,2)$ hand has 4 degrees of freedom according to the general mobility formula. Two of them correspond to the wrist rotations, while the other two are in-hand degrees of freedom. This allows us to include the ability of the fingers to be compatible with a contact constraint at a given position in the task.

It is possible to state the synthesis problem with as many velocities as desired, as long as they are compatible with the conditions in Eqs. (\ref{eq:mob}) and (\ref{eq:position}).

\section{Kinematic Solver}
\IEEEPARstart{T}{he} algorithms presented in this work have been implemented in kinematic design software. A first version of the solver for dimensional synthesis, \emph{ArtTreeKS} (Articulated Tree Kinematic Synthesis), was developed \cite{Simo-Serra:2011} for tree topologies with a single branching, corresponding to anthropomorphic or simple hands. A single root is sought for the system of equations using a hybrid solved based on a Genetic Algorithm (GA) built on top of a Levenberg-Marquadt local optimizer, which minimizes the average error of the dual quaternions representing the task. This numerical solver yields a single solution but allows dealing with tree topologies with a very high number of joints and fingers.   

As all meta-heuristic algorithms, this genetic algorithm must be adjusted experimentally according to the problem being solved.  Each entity in the genetic algorithm is represented as a vector of real numbers that allows simple integration with other numerical solver libraries like MINPACK \cite{More:1980}, which is used in the Hybrid solver.

This numerical solver has been integrated into the kinematic design package. The package includes a type synthesis stage, solvability checking, the ability to synthesize new designs with arbitrary branching stages (corresponding to hands with several palms), and the ability to define a task with positions and several velocities or accelerations at a given position. This second feature is important to fully determine manipulation actions such as finger rolling or finger sliding, or simple dexterity without changing the grasping point.

\section{Overall Software Architecture}
\begin{figure}[h]
\centering
\includegraphics[width=3.7in]{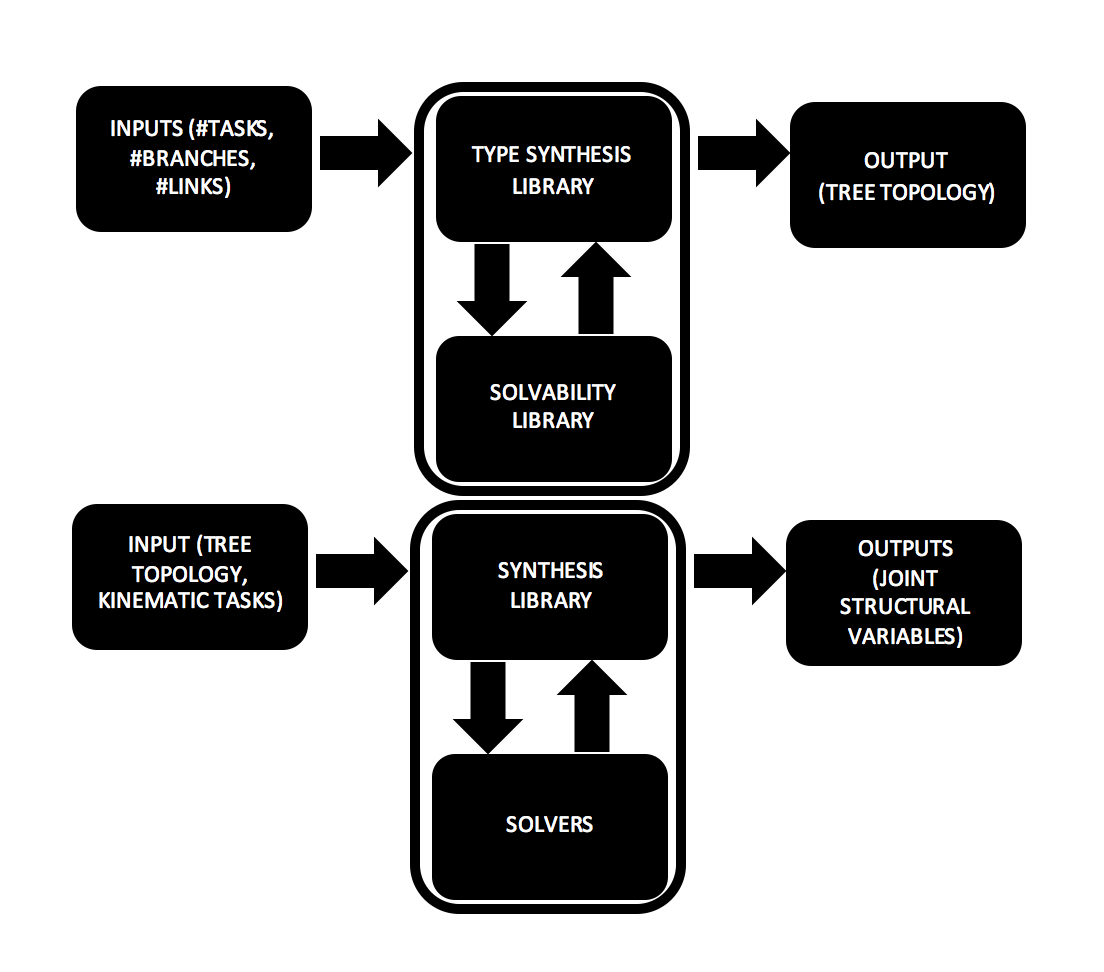}
\caption{ Software architecture.}
\label{fig:flowchart2}
\end{figure}

\IEEEPARstart{T}{he} software implementation follows a three-layer architecture, which is shown in Figure \ref{fig:flowchart2}, and uses the elements described below. The user interface and input files are developed using Lua, while the solver is programmed using C++.

\renewcommand{\labelitemii}{$\cdot$}
\begin{itemize}
\item Input Files: Input information from the designer. 
\begin{itemize}
\item Type synthesis input: contains the number of branches, task positions, and edges.
\item Dimensional synthesis input: Contains the tree topology and the values for the task positions, velocities, and accelerations.
\end{itemize}
\item Output files: Results of calculations that the designer can access.
\begin{itemize}
\item Solvability output: output file with the results of solvability calculations for a given topology.
\item Type synthesis output: output file with the atlas of solvable topologies for a given number of fingers and positions.
\item Design Synthesis File: output file containing the design equations.
\item Dimensional synthesis output: output file containing the result of the dimensional synthesis: Plucker coordinates of joint axes and joint variables.
\end{itemize}
\item Process files: internal calculations
\begin{itemize}
\item Solvability Library: Lua functions to calculate tree solvability.
\item Type synthesis Library: Lua file to construct all possible topologies for a set of input conditions.
\item Generator File: Lua functions to check solvability, assemble forward kinematics equations, and assign initial values.
\item Synthesis Library: Library of functions to communicate solver and Synthesis file.
\item Solvers: Genetic algorithm and Minpack C++ code to generate candidate solutions and perform minimization.
\end{itemize}
\end{itemize}

\section{Design Example}
\IEEEPARstart{A}{s}As an illustration of the overall design process, let us consider a hand task that can be defined with five positions of each fingertip. We can use a minimum of three fingers and a maximum of five fingers for this task for grasping and manipulation purposes. Three fingers may be sufficient for stable grasping, but adding the extra two fingers may help in some manipulation strategies.

We start the design process defining $m_p=5$ number of task positions, three to five branches $b=3$ to $b=5$, and we limit the number of edges to the interval from $e=1$ to $e=9$ in order to have a bounded search and to limit the complexity of the design. 

Applying Algorithms \ref{alg:search} and \ref{alg:solv}, we find, for $b=3$, a total of 29 solvable non-isomorphic topologies. For $b=4$ there are 134 solvable topologies, and for $b=5$ we find a total of 728 solvable topologies. For $b=4$ and $b=5$ fingertips, we notice that the minimum number of edges for the solvable topologies is $e=4$ and $e=5$ respectively. In order not to complicate the design too much, we limit the search to a maximum of $e=5$ number of edges. Table \ref{tab:solvTopEx} shows the solutions of the type synthesis stage for 3 to 5 fingertips and up to 5 edges. 

\begin{table}[h]
\caption{Solvable topologies for 5 positions, 3 to 5 fingertips}
\label{tab:solvTopEx}
\begin{center}
\begin{tabular}{|c|c|c|c|c|}
\hline
\textbf{Fingers} & \textbf{Edges} & \textbf{Topologies} & \textbf{Parent-pointer} & \textbf{Joint} \\
\hline \hline
 & $e=3$ & 1& $\{0,0,0\}$&  $\{3,3,3\}$\\
\cline{2-5}
  &  &  &  & $\{1,3,2,3\}$ \\
  & & &$\{0,0,1,1\}$ & $\{2,3,1,3\}$ \\
  & & & & $\{2,3,2,2\}$ \\  
  & & & & $\{3,3,1,2\}$ \\ \cline{4-5}  
   &$e=4$ &9 &  & $\{1,2,3,3\}$\\
   & & & & $\{2,1,3,3\}$ \\
   & & &$\{0,1,1,1\}$ & $\{2,2,2,3\}$ \\
  & & & & $\{3,1,2,3\}$ \\
  & & & & $\{3,2,2,2\}$ \\
 \cline{2-5}
  & &  &  & $\{1,1,2,2,3\}$ \\
  & & & & $\{1,1,3,1,3\}$ \\
  & & & & $\{1,1,3,2,2\}$ \\
  & & & & $\{1,2,2,1,3\}$ \\
  $b=3$& & & & $\{1,2,2,2,2\}$ \\
  & & & & $\{1,2,3,1,2\}$ \\
  & & & & $\{1,3,2,1,2\}$ \\
  & & & & $\{2,1,1,2,3\}$ \\
  & & & & $\{2,1,2,1,3\}$ \\
  & $e=5$ & 19 &$\{0,1,1,2,2\}$ & $\{2,1,2,2,2\}$ \\
  & & & & $\{2,1,3,1,2\}$ \\
  & & & & $\{2,2,1,1,3\}$ \\
  & & & & $\{2,2,1,2,2\}$ \\
  & & & & $\{2,2,2,1,2\}$ \\
  & & & & $\{2,3,1,1,2\}$ \\
  & & & & $\{3,1,1,1,3\}$ \\
  & & & & $\{3,1,1,2,2\}$ \\
  & & & & $\{3,1,2,1,2\}$ \\
  & & & & $\{3,2,1,1,2\}$ \\
\hline
 & $e=4$ & 1 & $\{0,0,0,0\}$ &  $\{3,3,3,3\}$\\
\cline{2-5}
  &  & &  & $\{1,3,3,2,3\}$ \\
    & & & $\{0,0,0,1,1\}$& $\{2,3,3,1,3\}$ \\
   & & & & $\{2,3,3,2,2\}$ \\
   & & & & $\{3,3,3,1,2\}$ \\\cline{4-5}
   & & &  & $\{1,3,2,3,3\}$ \\   
   & & & & $\{2,3,1,3,3\}$ \\
$b=4$ &$e=5$ &14 & & $\{ 2,3,2,2,3\}$ \\
   & & & & $\{3,3,1,2,3\}$ \\
   & & & & $\{3,3,2,2,2\}$ \\
   & & & $\{0,0,1,1,1\}$& $\{1,2,3,3,3\}$ \\
   & & & & $\{2,1,3,3,3\}$ \\
   & & & & $\{2,2,2,3,3\}$ \\
   & & & & $\{3,1,2,3,3\}$ \\
   & & & & $\{3,2,2,2,3\}$ \\
\hline
$b=5$ & $e=5$ & 1 & $\{0,0,0,0,0\}$ &  $\{3,3,3,3,3\}$\\
\hline
\end{tabular}
\end{center}
\end{table}


The simplest solvable topology able to perform this task has parent-pointer array $p=\{0,0,0\}$ and joint array $j=\{3,3,3\}$, that is, a $0-(3,3,3)$ topology with three R-R-R fingers and no wrist. Some of the most complex topologies are the $3-(2,1-(1,2))$ topology, with $p=\{0,1,1,2,2\}$ and $j=\{3,2,1,1,2\}$, or the $0-(3,2-(2,2,3))$, with $p=\{0,0,1,1,1\}$ and $j=\{2,3,2,2,3\}$. 

\begin{figure}[h]
\centering
\includegraphics[width=2.6in]{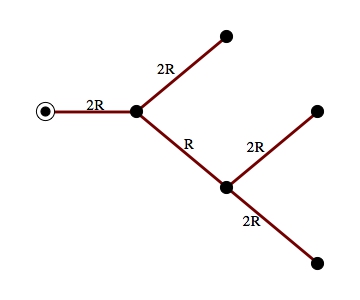}
\includegraphics[width=2.8in]{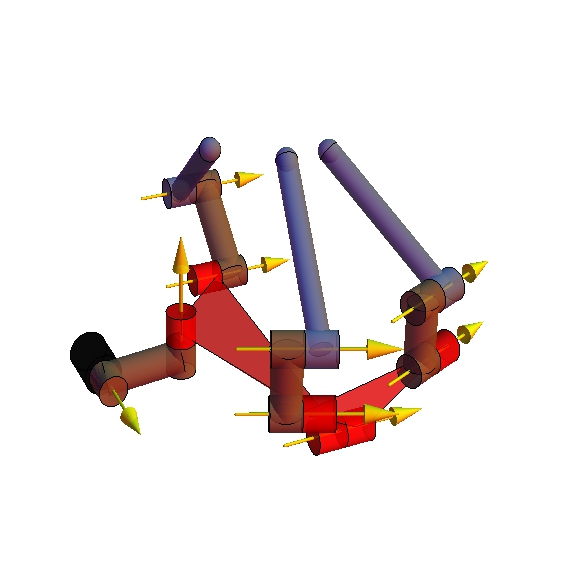}
\caption{ Graph of the selected topology and its kinematic sketch.}
\label{fig:example01122}
\end{figure}

Out of the 45 candidate topologies, we select for the design the topology with $p=\{0,1,1,2,2\}$ and $j=\{2,1,2,2,2\}$, corresponding to the $2-(1-(2,2),2)$ hand, with two R joints at the wrist spanning two fingers, the first one spanning two more fingers for a total of three end-effectors. Figure \ref{fig:example01122} shows the graph of this topology.
   
Dimensional synthesis is used to shape this topology with five random finite displacements. The resulting set of equations from Eq.(\ref{eq:treedesign}) consists of 96 highly nonlinear equations in 90 unknowns, 72 of which are independent. The results of five runs with different initial conditions are presented in Table \ref{tab:runs}. All five obtained solutions were feasible and different, which makes us infer that there will be a large number of solutions for this topology. 

\begin{table}[h]
\caption{Dimensional synthesis solver results}
\label{tab:runs}
\begin{center}
\begin{tabular}{|lc|c|c|c||}
\hline
\textbf{Run} & \textbf{Final error} & \textbf{Iterations} & \textbf{Running time} \\
\hline \hline
1 & $1.15*10^{-13}$ & 1 & 12 sec. \\
2 & $6.0*10^{-12}$ & 2 & 10 sec. \\
3 & $1.4*10^{-13}$ & 1 & 5 sec. \\
4 & $2.0*10^{-13}$ & 9 & 29 sec. \\
5 & $1.0*10^{-13}$ & 1 & 7 sec. \\
\hline
\end{tabular}
\end{center}
\end{table}

\begin{figure}[h]
\centering
\includegraphics[width=3.0in]{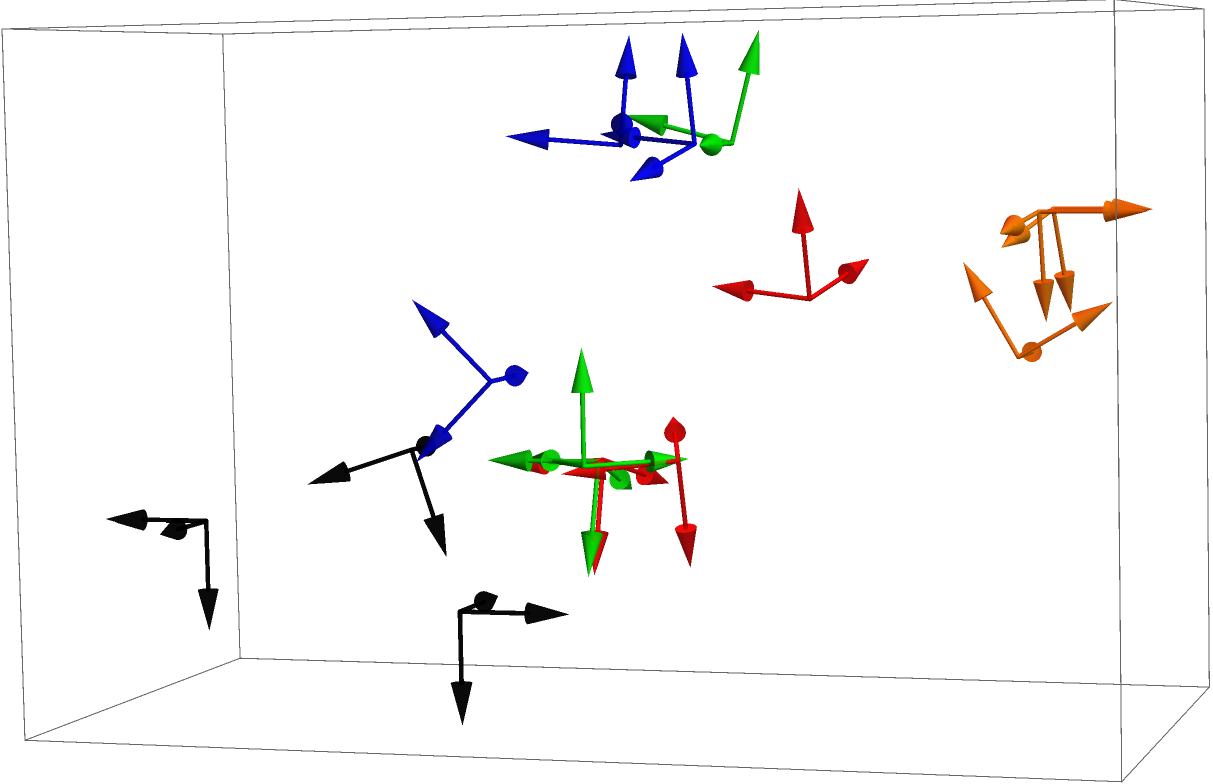}
\caption{The position task. Colors correspond to each position of all three fingertips.}
\label{fig:positionsExample}
\end{figure}

\begin{figure}[h]
\centering
\includegraphics[width=3.0in]{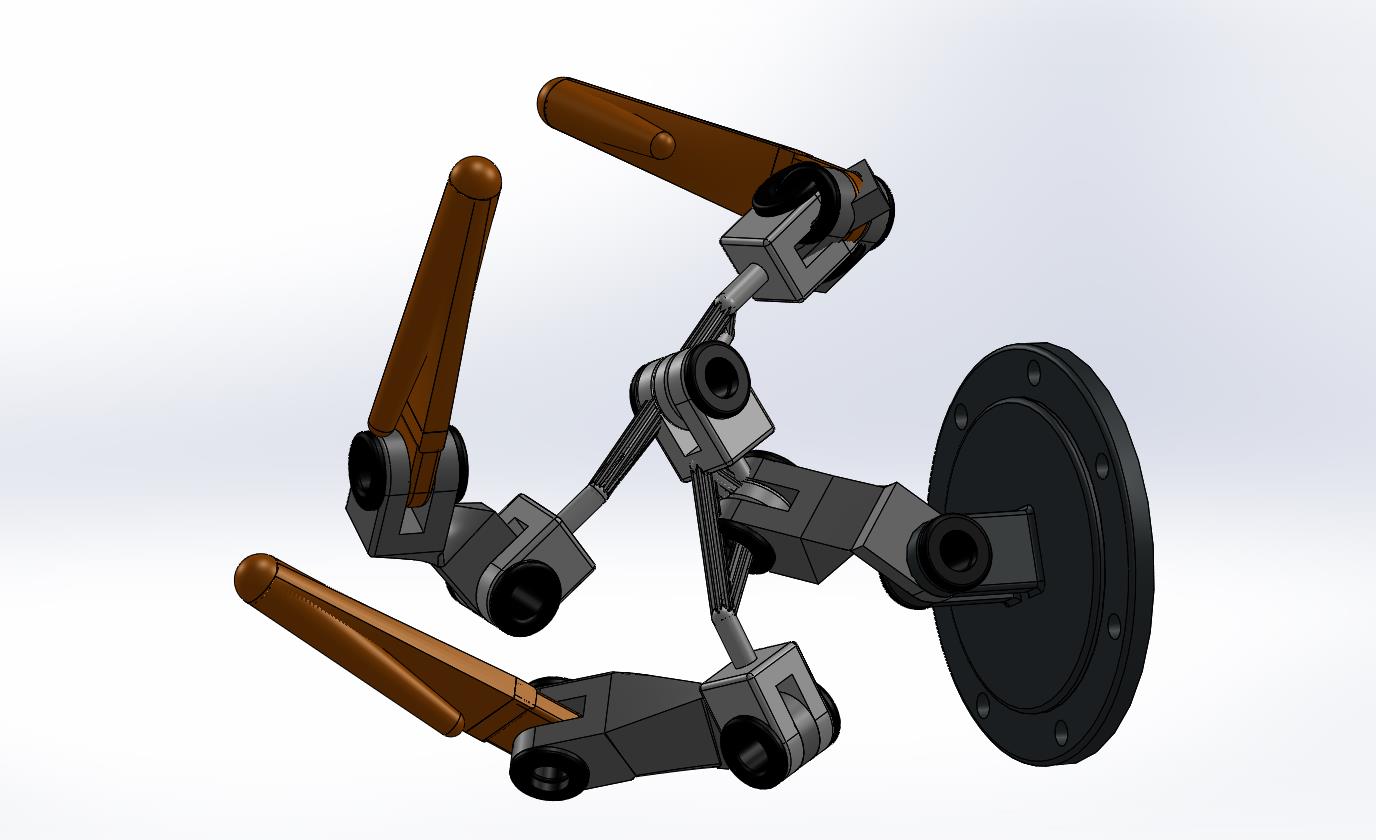}
\caption{Hand design using the topology 2-(1-(2,2),2)}
\label{fig:drawingExample}
\end{figure}

\begin{figure}[h]
\centering
\includegraphics[width=3.5in]{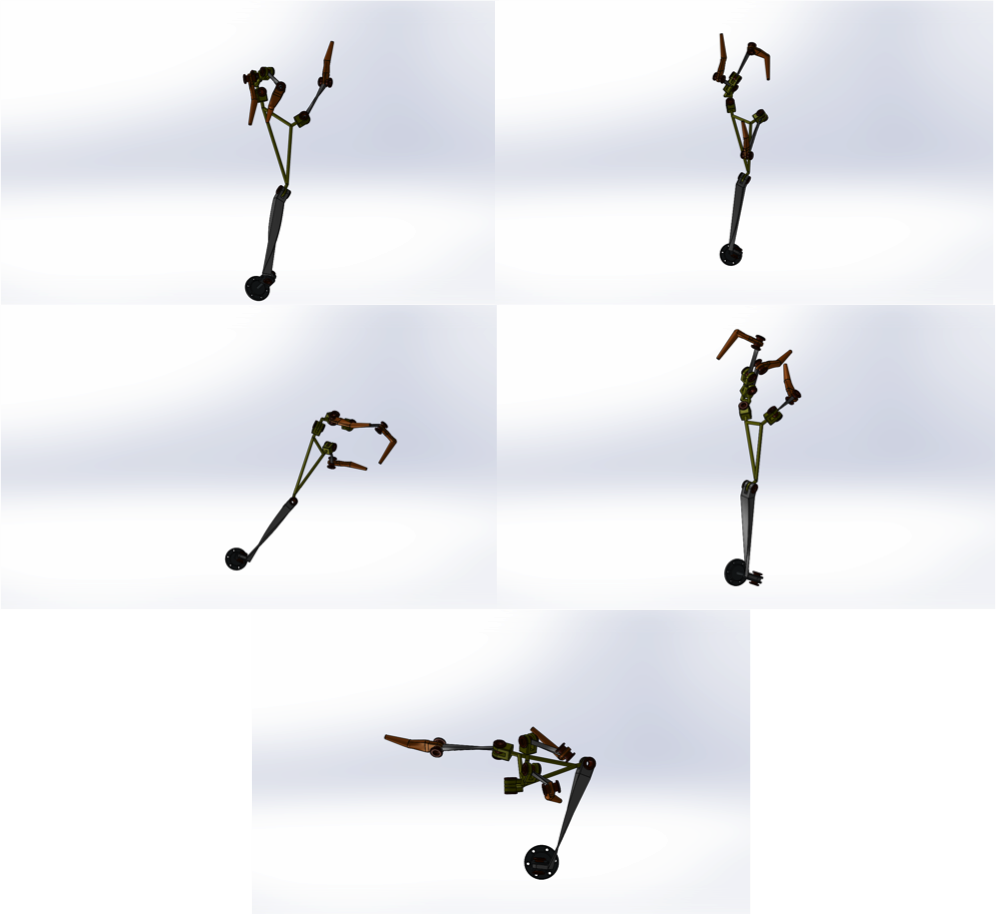}
\caption{2-(1-(2,2),2) hand design for the specified positions.}
\label{fig:configs}
\end{figure}

The solutions obtained with the dimensional synthesis solver have been modeled using the automatic drawing procedure developed in \cite{Hassanzadeh:2015} and are presented in Figure \ref{fig:drawingExample} and Figure \ref{fig:configs}. The positions used for this design are presented in Figure \ref{fig:positionsExample}. 

An example of a multi-ﬁngered robotic hand with ﬁve RRSS close chains and a common R joint at the wrist is presented in this section.  This hand, as a member of the single-jointed tree topology family design process, was analyzed in \cite{Makhal:2014},\cite{Deemyad:2018}, and dimensional synthesis of it was studied in \cite{Perez:2015}. Another five-finger hand with a similar idea and using five Bennett linkages and a common single revolute joint in the wrist was analyzed in \cite{Deemyad:2016}. Synthesis methodology steps for the 1 (1, 1, 1, 1, 1) hand contains five fingers with a single revolute joint which are connected to a common single revolute joint in the wrist by a single palm is completely discussed in \cite{SimoSerra:2014}.  For designing a hand with five RRSS close chains only need to couple each of RR chain with an SS chain. On the other hand, the SS chain have to reach the same task positions in which RR was synthesized based on them. Therefore, Equation \ref{eq:ss-1} must satisfy where $\hat{P_{1_2}}$ is the relative position between task pos1 and pos2. In this equation, $\alpha$ and $\beta$  are S joints axes-angles.

\begin{equation}
\label{eq:ss-1} 
\hat{S}_2(\beta)=\hat{S}_1^*(\alpha)\hat{P}_{1_2}
\end{equation}

By using the inverse kinematics for the axes-angles $\alpha$ and $\beta$, an $8 \times 8$ matrix system will be found. To avoid a nonzero solution for the rotation variables, this matrix has to be not a full rank. This matrix and all other steps are mentioned in detail in \cite{Deemyad:2018}. Finally, this problem can be solved as an exact synthesis problem by selecting six of the seven parameters and finding the last one or continue as an optimization problem to find the minimum size for this hand when having more control on the other requirements. 

After coupling SS chains to the fingers, Mathematica and Matlab Global Optimization toolbox, genetic algorithm method, and fmincon have been applied to reach the most optimized hand with minimum size and several other limitations. According to the final goal for this design, which is minimizing the hand’s total size, the objective function is defined as the summation of the lengths between all four joints, (R, S, S, R joints) of all five fingers. Also, for reaching an applicable final design for this hand,  six various sets of equality and non-equality nonlinear constraints were used for this optimization. Four of these constraints were non-equality nonlinear constraints: 1) The minimum and maximum range for each link’s length. 2) A cylindrical space for each joint motion from first to second position has been defined to avoid the self-intersection. 3) The range of angular rotation (max 1.8 Radian) for each S joint has been limited based on available and practical S joint in industry. 4) Some constraints have been defined to avoid passing the singularity points during the hand motion from first to the second position for all fingers, which is presented in \cite{Deemyad:2020}. Also, two remaining constraints were equality nonlinear constraints: 5) The determinant of the $8 \times 8$ matrix system for each finger, which was defined before have to be equal zero. 6) The loop closure equation for each finger has to be satisfied. Each finger was optimized separately in a time-consuming process. For each of them, it takes at most 200 seconds and terminated at less than 200 generations.

\begin{figure}[h]
\centering
\includegraphics[width=3.0in]{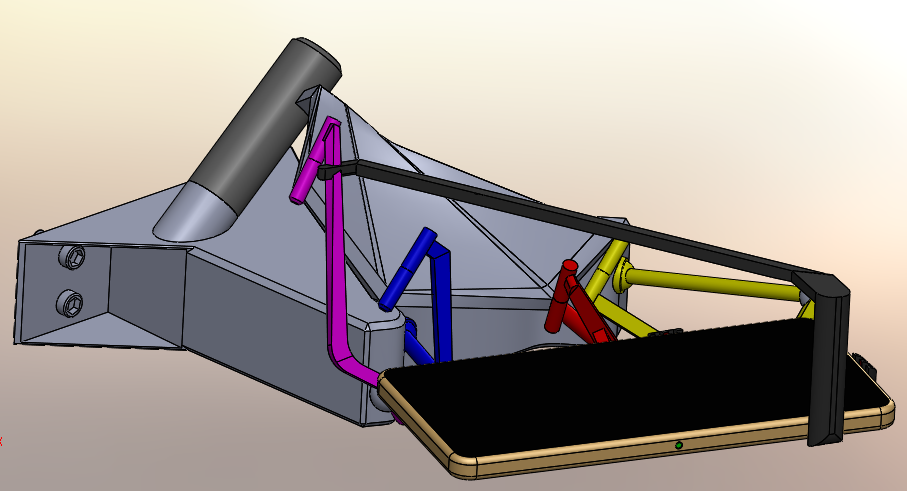}
\caption{Hand with five RRSS chains (1st view).}
\label{fig:RRSSExample1}
\end{figure}

\begin{figure}[h]
\centering
\includegraphics[width=3.0in]{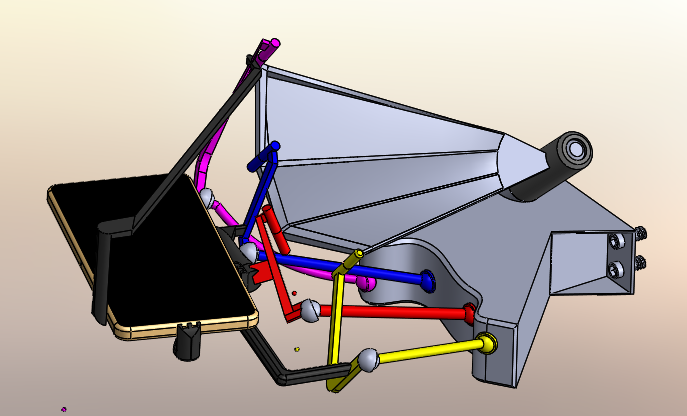}
\caption{Hand with five RRSS chains (2nd view).}
\label{fig:RRSSExample2}
\end{figure}

Different views for the final optimized model of this hand with five RRSS closed chains are shown in Figures \ref{fig:RRSSExample1} and \ref{fig:RRSSExample2} which was designed and simulated in SolidWorks. This hand just by using one actuator grasps a mobile phone and simultaneously presses a touch button.

The output of the kinematic synthesis stage is to be used as the input for a detailed design, using computer-aided tools.

\section{Conclusions}

This work presents the development and implementation of the different kinematic design stages within a tool for the creation of innovative multi-fingered robotic hands. The resulting design package can perform type and dimensional kinematic synthesis for arbitrary tree topologies, enumerating candidate topologies, and creating wristed hands with arbitrary number and type of fingers and arbitrary number and type of branchings. The solver accepts several inputs, basically a kinematic task and some limits on the desired topologies such as the number of fingers or some bounds on the edges. The kinematic task may include finite displacements of each fingertip and multiple velocities and accelerations for the fingertips at some of those finite positions. 

Type synthesis and solvability are implemented using an enumeration technique that constructs non-isomorphic trees. The dimensional synthesis's implementation combines the automatic construction of the tree forward kinematics with a solver consisting of a genetic algorithm and a Levenberg-Marquardt stage to explore the space of solutions and shows fast convergence to a solution for each run. The current version of the solver is freely available on the project webpage.  Future work will focus on generalizing some other solver features and on the automatic connection to subsequent stages in the design process.

The design process's output is a \emph{kinematic design}: a set of joint axes, defined by their Plucker coordinates at a reference configuration, and a set of joint variables and joint rates. Each kinematic design can be implemented in the final design in an unlimited number of ways, selected by the designer and constrained by additional specifications.  The rationale is that a hand design tailored to an application may simplify many other aspects of the process, increasing the grasping and manipulation actions' success.






\ifCLASSOPTIONcaptionsoff
  \newpage
\fi



%


\bibliographystyle{plainnat}
\bibliography{NSF-RHref}



\end{document}